\documentclass[10pt,twocolumn,letterpaper]{article}

\usepackage[pagenumbers]{cvpr} 

\usepackage{array}
\usepackage{colortbl}
\usepackage{booktabs}
\usepackage{tabularx}

\usepackage{multirow}

\newcommand{\piZero}{\ensuremath{\pi_{0}}\xspace}
\newcommand{\piHalf}{\ensuremath{\pi_{0.5}}\xspace}
\newcommand{\best}[1]{\textbf{#1}}
\newcommand{\second}[1]{\underline{#1}}
\newcommand{\cmark}{\ensuremath{\checkmark}}
\newcommand{\xmark}{\ensuremath{\times}}

\definecolor{oursbg}{RGB}{240,240,240}

\makeatletter
\apptocmd{\@maketitle}{%
    \par\noindent
    \begin{minipage}{\textwidth}
        \centering
        \includegraphics[
            pagebox=cropbox,
            width=\linewidth
        ]{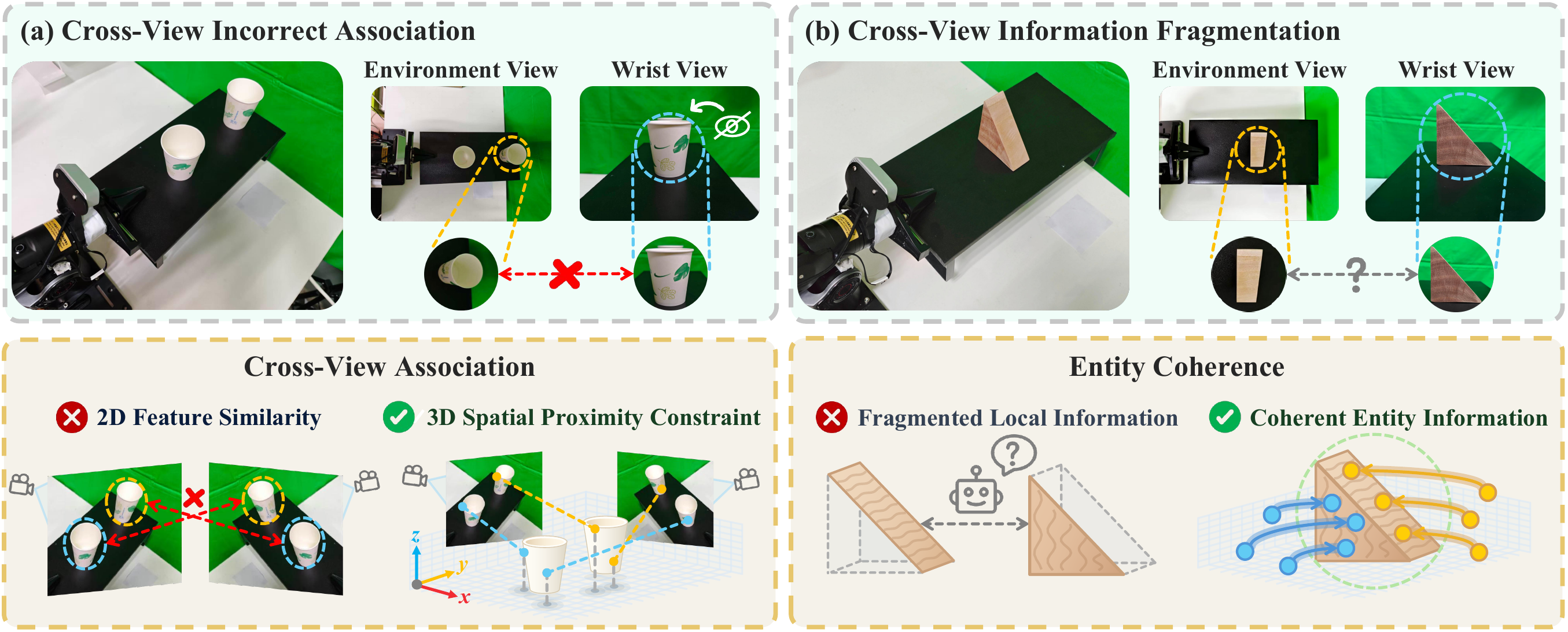}
        \captionsetup{hypcap=false}
        \captionof{figure}{Motivation of ConsisX. (a) Visual similarity between distinct physical entities may lead to incorrect cross-view association. ConsisX introduces 3D spatial proximity constraint for cross-view association. (b) Variations in local observations of the same physical entity may lead to cross-view information fragmentation. ConsisX organizes complementary 3D information into coherent entity evidence.}
        \label{fig:motivation}
    \end{minipage}
    \par\vspace{10pt}
}{}{}
\makeatother

\definecolor{cvprblue}{rgb}{0.21,0.49,0.74}
\usepackage[pagebackref,breaklinks,colorlinks,allcolors=cvprblue]{hyperref}

\def\confName{CVPR}
\def\confYear{2027}

\title{ConsisX: Cross-View Consistency Thinking in 3D for Vision-Language-Action Models}

\author{Yunhe Zhang$^{1}$, Long Gao$^{1}$, Yuedong Fu$^{1}$, Gang He$^{1}$, Li Xu$^{1}$, Chang Wu$^{1}$, Yuqi Zhu$^{1}$, Pengyu Li$^{2}$\\
$^{1}$Xidian University \quad $^{2}$Beihang University\\
{\tt\small yunhezhang@stu.xidian.edu.cn}
}

\begin{document}
\maketitle
\begin{abstract}
Vision-language-action (VLA) models integrate visual perception and language understanding from vision-language models (VLMs) with action prediction for robotic manipulation. Although multi-view observations provide complementary scene information, VLMs model cross-view interactions based on 2D feature similarity, which can lead to physical inconsistency in the form of cross-view incorrect association between distinct entities and cross-view information fragmentation about the same entity. To address these challenges, we propose \textbf{ConsisX}, a novel VLA framework that integrates cross-view association and entity coherence to enable cross-view consistency thinking in 3D for physically consistent robotic manipulation. Specifically, Cross-View Association Modeling (CAM) is proposed to introduce cross-view spatial proximity constraints in a shared 3D space to reduce cross-view incorrect association. Entity Coherence Organization (ECO) is designed to organize cross-view local information into coherent 3D entity evidence, thereby addressing cross-view information fragmentation. Consistent Action Conditioning (CAC) is introduced to condition action prediction on task-relevant 3D entity evidence for cross-view consistent action execution. Experiments on 3 real-world manipulation tasks and 4 simulation benchmarks establish ConsisX as an effective and competitive framework for robotic manipulation.

\end{abstract}
\section{Introduction}
\label{sec:intro}

In modern embodied intelligence systems, robots take natural language instructions and multi-view visual observations as input to predict actions for task execution. Recently, vision-language-action (VLA) models have advanced robotic manipulation by extending the visual and language understanding of pre-trained vision-language models (VLMs) \cite{achiam2023gpt, beyer2024paligemma} to robot control through an action prediction module optimized via demonstration learning \cite{wang2026vla, zheng2026x}. RT-2 and OpenVLA adapt VLMs into robotic policies to predict actions as discrete tokens \cite{zitkovich2023rt, kim2024openvla}. Subsequently, \piZero introduces an action expert to predict continuous action sequences \cite{black2024pi_0}. Building upon the \piZero, \piHalf combines diverse robot and vision-language data during training for enhanced generalization to unseen environments \cite{intelligence2025pi_}.

A common setup for robotic policies incorporates multi-view visual inputs from environment and wrist cameras to provide complementary observations of the current scene for reliable action prediction. To integrate these observations, VLA models enable interactions between visual features from different views through jointly processing within the VLM. However, the joint processing can introduce cross-view ambiguity. As shown in \cref{fig:motivation}(a), both cups appear in the environment view but only one is visible in the wrist view, and feature interactions may link the front one to the rear one. In \cref{fig:motivation}(b), the two cameras capture different faces of the same triangular prism, and 2D feature similarity alone may fail to link the local observations.


The above cases highlight a challenge: \textit{cross-view interactions in VLMs rely on learned similarities between 2D features and can therefore lead to physically inconsistent action execution.} Specifically, the challenge involves two forms of cross-view ambiguity. \textbf{(Q1) Visually similar observations may be incorrectly associated despite corresponding to distinct physical entities. (Q2) Cross-view information remains fragmented despite coming from the same physical entity.}



Do visually similar observations across views actually refer to the same physical entity? How should fragmented information from different views of the same physical entity be organized? We regard these ambiguities as a 3D cross-view consistency problem and argue that cross-view thinking should jointly consider \textbf{cross-view association} and \textbf{entity coherence} in 3D. Specifically, visual similarity alone is insufficient to establish physical identity, and cross-view association is therefore needed to provide 3D spatial constraints. The fragmented information from the same entity also requires entity coherence to guide its organization into physically consistent evidence. The above considerations lead to the central question: \textit{How to think about cross-view association and entity coherence to enable cross-view consistency in 3D?}

To answer this question, we propose ConsisX, a novel VLA framework that combines cross-view association and entity coherence for cross-view consistency thinking in 3D. Specifically, the consistency thinking follows a progressive process. Firstly, \textit{Cross-View Association Modeling (CAM)} is proposed to organize 3D features from each view in a shared 3D coordinate system and infer cross-view association with spatial proximity constraints and feature similarity. \textbf{(A1) CAM provides cross-view correspondences under 3D spatial constraint, thereby reducing cross-view incorrect association.} Secondly, \textit{Entity Coherence Organization (ECO)} groups fragmented local 3D feature within each view into entity candidates through spatial affinity and feature compatibility before organizing cross-view entity pairs according to the cross-view association from CAM. \textbf{(A2) ECO organizes cross-view local information into coherent 3D entity evidence, thus addressing cross-view information fragmentation.} Finally, \textit{Consistent Action Conditioning (CAC)} retrieves candidate entity pairs from ECO according to task relevance, providing the corresponding 3D entity features for cross-view consistent action prediction. ConsisX thus jointly considers cross-view association and entity coherence, enabling cross-view consistency thinking in 3D for reliable robotic manipulation. The main contributions of this work can be summarized as follows.
\begin{itemize}
    \item To address the challenge of cross-view physical inconsistency, we propose ConsisX, a novel VLA framework that integrates cross-view association and entity coherence to enable cross-view consistency thinking in 3D.
    \item We design CAM to reduce cross-view incorrect associations through 3D spatial constraint, ECO to organize fragmented local information into coherent 3D entity evidence, and CAC to provide task relevant 3D entity features for cross-view consistent action prediction.
    \item We evaluate ConsisX on 4 simulation benchmarks and conduct real-world experiments on 3 manipulation tasks, demonstrating the effectiveness and competitive performance of the proposed framework.
\end{itemize}

\section{Related Work}
\label{sec:formatting}

\subsection{Vision-Language-Action Models}

VLA models integrate visual perception, language understanding, and action prediction for robotic manipulation \cite{kim2024openvla, black2024pi_0, intelligence2025pi_}. Recent progress builds on scalable robot learning and pre-trained vision language knowledge. RT-1 \cite{brohan2022rt} and Octo \cite{team2024octo} learn from diverse robot demonstrations, while RT-2 \cite{zitkovich2023rt} and OpenVLA \cite{kim2024openvla} adapt pre-trained VLMs. To further advance VLA models, some studies explore ways to improve policy adaptation and generalization. 
OpenVLA-OFT \cite{kim2025fine} improves adaptation with optimized fine-tuning, and \piHalf \cite{intelligence2025pi_} enhances generalization through more diverse data. For action prediction, FAST \cite{pertsch2025fast} compresses action sequences into discrete tokens, and \piZero \cite{black2024pi_0} introduces flow matching for continuous action prediction. Beyond direct action prediction, some methods integrate intermediate reasoning to improve action prediction \cite{zawalski2024robotic, zhao2025cot, zhong2026acot}.

\begin{figure*}[t]
    \centering
    \includegraphics[width=\textwidth]{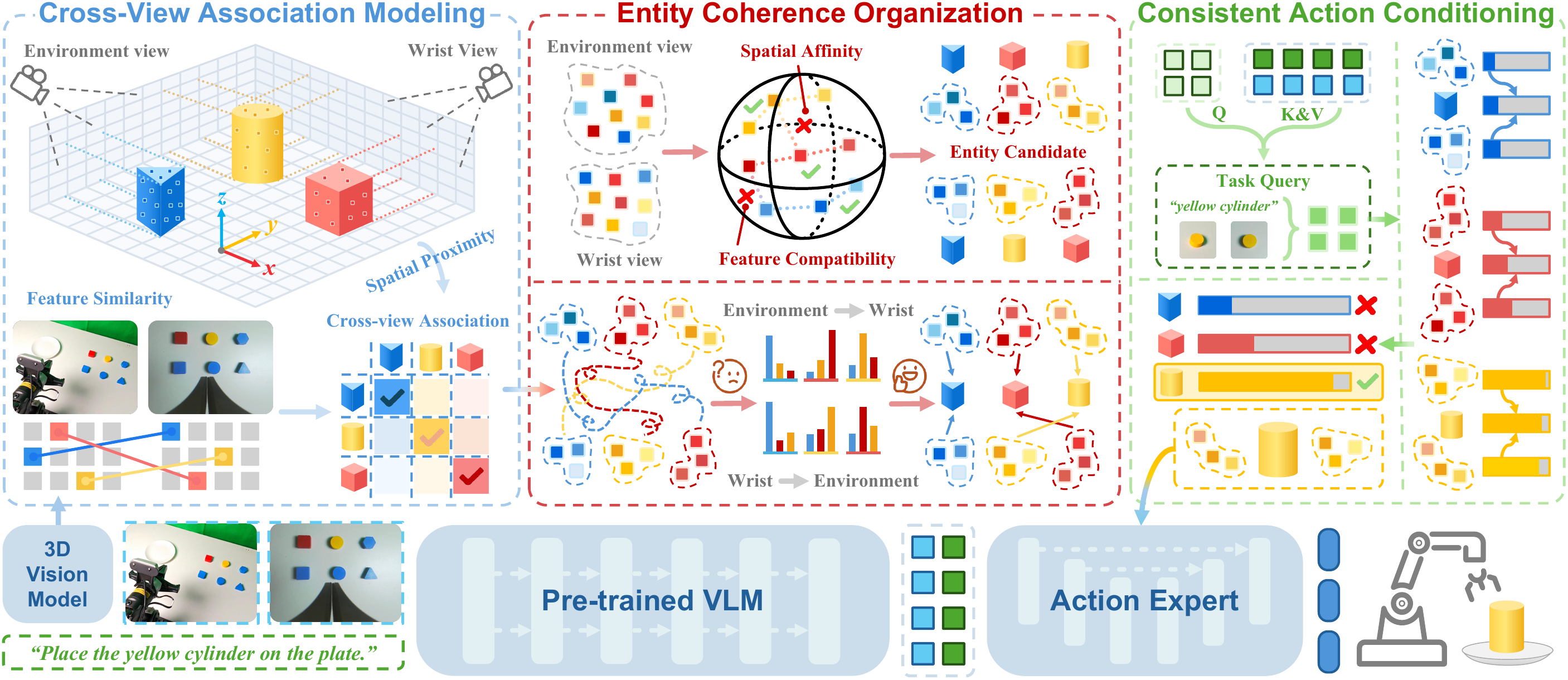}
    \caption{Overview of ConsisX. Cross-View Association Modeling (CAM) introduces 3D spatial proximity constraint to reduce cross-view incorrect association. Entity Coherence Organization (ECO) employs coherent entity modeling to address cross-view information fragmentation. Consistent Action Conditioning (CAC) provides task-relevant 3D entity evidence for physically consistent action prediction.}
    \label{fig:pipeline}
\end{figure*}

\subsection{Cross-View Association Learning}

Cross-view association learning establishes correspondences between different views for consistent scene understanding. In computer vision, SuperGlue \cite{sarlin2020superglue} relies on cross-view feature similarity, and CCMP \cite{yan2026learning} identifies corresponding object instances. Most VLA policies integrate multiple views through joint feature interactions within the VLM \cite{black2024pi_0, intelligence2025pi_}. StereoVLA \cite{deng2025stereovla} further integrates cross-view information through stereo matching, and ConsisVLA-4D \cite{li2026consisvla} enhances task-relevant cross-view features with geometric priors. Moreover, XVR \cite{jeong2026learning} optimizes VLMs to improve cross-view understanding in VLA policies.

\subsection{3D-Aware Policy Learning}

3D-aware learning improves spatial understanding in robotic policies for reliable action execution in the physical world. Some approaches incorporate additional spatial information into policy learning. PerAct \cite{shridhar2023perceiver} predicts actions from voxelized RGB-D observations, and RVT \cite{goyal2023rvt} aggregates point cloud-based virtual views. For VLA models, 4D-VLA \cite{zhang20264d} augments visual feature with 3D position encoding, and PointACT \cite{chen2026pointact} introduces point cloud feature for action prediction. Researchers further explore 3D object keypoint constraints \cite{li20253ds} and 3D-consistent waypoints \cite{huang20263dway} for robot control. Other VLA methods leverage pre-trained 3D vision models to refine visual features \cite{feng2026spatial, li2026spatial}.

In contrast, ConsisX integrates cross-view association and entity coherence to enable cross-view consistency thinking in 3D. Through spatially constrained cross-view association and coherent entity organization, ConsisX leverages cross-view 3D entity evidence for physically consistent action prediction.

\section{Method}

This section details how ConsisX enables cross-view consistency thinking in 3D. \cref{sec:method_preliminary} introduces the problem formulation and the base policy. 
We then detail CAM in \cref{sec:method_cam}, ECO in \cref{sec:method_eco}, and CAC in \cref{sec:method_cac}. The training and inference are finally described in \cref{sec:method_other}.

\subsection{Preliminary}
\label{sec:method_preliminary}

\textbf{Problem formulation.} Given a language instruction $\ell$ and multi-view visual observations $\mathcal{O}_t = \{I_t^{E}, I_t^{W}\}$ from an environment camera and a wrist camera at time $t$, the policy $\pi_{\theta}$ predicts the next $H$ continuous actions as an action chunk $\mathbf{A}_t = [\mathbf{a}_t, \mathbf{a}_{t+1}, \ldots, \mathbf{a}_{t+H-1}]$. The process can be expressed as follows.
\begin{equation}
    \pi_{\theta}: (\mathcal{O}_t, \ell) \rightarrow \mathbf{A}_t.
    \label{eq:policy}
\end{equation}

\noindent\textbf{Base policy.} ConsisX is built on \piHalf \cite{intelligence2025pi_}, a base policy that combines a pre-trained VLM \cite{beyer2024paligemma} with a flow-matching action expert \cite{lipman2022flow}. The VLM jointly processes the current multi-view visual observations and the language instruction embeddings to produce layer-wise vision-language features. The action expert is conditioned on these layer-wise features and iteratively refines a noisy action sequence to produce the final action chunk.

In the base policy, cross-view interactions within the VLM rely on 2D visual feature similarity and may lead to physical inconsistency in the form of cross-view incorrect association between distinct entities and cross-view information fragmentation about the same entity. ConsisX therefore introduces CAM, ECO, and CAC as a progressive process for cross-view consistency thinking in 3D.

\subsection{Cross-View Association Modeling}
\label{sec:method_cam}

To reduce cross-view incorrect association, CAM is proposed to establish matches between 3D features from different views by incorporating spatial proximity constraint in a shared 3D coordinate system with feature similarity.

\noindent\textbf{Cross-View Shared Coordinate.} For 3D feature construction, a pre-trained 3D vision model jointly processes the environment and wrist observations to produce their respective 3D features $\mathbf{F}^{E} = \{\mathbf{f}_{1}^{E}, \mathbf{f}_{2}^{E}, \ldots, \mathbf{f}_{N}^{E}\}$ and $\mathbf{F}^{W} = \{\mathbf{f}_{1}^{W}, \mathbf{f}_{2}^{W}, \ldots, \mathbf{f}_{N}^{W}\}$, along with depth estimates and camera parameters. For each view, the 3D feature is subsequently assigned coordinates in a shared 3D coordinate system through depth back-projection and camera coordinate transformation. The process of the environment view can be summarized as follows.
\begin{equation}
\begin{aligned}
\mathbf{T}_{E}
&=
\begin{bmatrix}
\mathbf{R}_{E} & \mathbf{t}_{E} \\
\mathbf{0}^{\top} & 1
\end{bmatrix}, \\
\begin{bmatrix}
\mathbf{p}_{i}^{E} \\
1
\end{bmatrix}
&=
\mathbf{T}_{E}^{-1}
\begin{bmatrix}
d_{i}^{E}\mathbf{K}_{E}^{-1}\mathbf{u}_{i}^{E} \\
1
\end{bmatrix},
\end{aligned}
\label{eq:shared_coordinates}
\end{equation}
where $\mathbf{T}_{E}$ is the rigid transformation defined by rotation $\mathbf{R}_{E}$ and translation $\mathbf{t}_{E}$, with its inverse $\mathbf{T}_{E}^{-1}$ mapping the environment camera coordinates into the shared 3D coordinate system. The index $i$ identifies the spatial location in $\mathbf{F}^{E}$. $\mathbf{u}_{i}^{E}$, $d_{i}^{E}$, and $\mathbf{K}_{E}$ denote the homogeneous image coordinate, predicted depth, and estimated camera intrinsics, respectively. The above process yields the shared 3D coordinate $\mathbf{p}_{i}^{E}$ for the spatial location $i$ in the environment view, and a similar process yields $\mathbf{p}_{i}^{W}$ for the wrist view. To eliminate the influence of coordinate scale variations on spatial distance comparisons, $\mathbf{p}_{i}^{E}$ and $\mathbf{p}_{i}^{W}$ are jointly normalized through a shared centroid and root mean square radius, resulting in $\widetilde{\mathbf{p}}_{i}^{E} = (x_{i}^{E}, y_{i}^{E}, z_{i}^{E})^{\top}$ and $\widetilde{\mathbf{p}}_{i}^{W} = (x_{i}^{W}, y_{i}^{W}, z_{i}^{W})^{\top}$, respectively.

\noindent\textbf{Cross-view Association Inference.} To maintain physical consistency thinking, CAM jointly considers spatial proximity constraint and feature similarity during cross-view association inference. First, the spatial proximity constraint is defined in terms of the spatial distance $D_{ij}^{EW}$ between $\widetilde{\mathbf{p}}_{i}^{E}$ from the environment view and $\widetilde{\mathbf{p}}_{j}^{W}$ from the wrist view in the shared 3D coordinate system as follows.
\begin{equation}
\scalebox{0.93}{$\displaystyle
D_{ij}^{EW}
=
\sqrt{
(x_{i}^{E} - x_{j}^{W})^{2}
+
(y_{i}^{E} - y_{j}^{W})^{2}
+
(z_{i}^{E} - z_{j}^{W})^{2}
}.
$}
\label{eq:spatial_distance}
\end{equation}
Subsequently, the cross-view feature similarity $S_{ij}^{EW}$ of $\mathbf{f}_{i}^{E}$ and $\mathbf{f}_{j}^{W}$ is quantified as follows.
\begin{equation}
S_{ij}^{EW}
=
\frac{
\langle \mathbf{f}_{i}^{E}, \mathbf{f}_{j}^{W} \rangle
}{
\lVert \mathbf{f}_{i}^{E} \rVert
\lVert \mathbf{f}_{j}^{W} \rVert
}.
\label{eq:feature_similarity}
\end{equation}
Finally, the cross-view association matrices $\mathcal{A}^{EW}, \mathcal{A}^{WE} \in [0,1]^{N \times N}$ are inferred by integrating the above spatial proximity constraint and feature similarity. Specifically, the cross-view association $\mathcal{A}_{ij}^{EW}$ from the environment view to the wrist view is computed as follows.
\begin{equation}
\begin{aligned}
\mathcal{A}_{ij}^{EW}
&=
\frac{
\exp\left(-\lambda(D_{ij}^{EW})^{2} + S_{ij}^{EW}\right)
}{
\sum_{u=1}^{N}
\exp\left(-\lambda(D_{iu}^{EW})^{2} + S_{iu}^{EW}\right)
}
\\
&\quad \times
\frac{
\exp\left(-\lambda(D_{ij}^{EW})^{2} + S_{ij}^{EW}\right)
}{
\sum_{v=1}^{N}
\exp\left(-\lambda(D_{vj}^{EW})^{2} + S_{vj}^{EW}\right)
},
\end{aligned}
\label{eq:cross_view_association}
\end{equation}
where $\lambda > 0$ is a learnable coefficient for the quadratic spatial penalty.

Through the incorporation of spatial proximity constraint from CAM, ConsisX reduces cross-view incorrect association compared with relying on feature similarity alone.

\subsection{Entity Coherence Organization}
\label{sec:method_eco}

To address cross-view information fragmentation, ECO organizes fragmented local 3D features within each view into consistent entity candidates and forms cross-view entity pairs by aggregating the cross-view association from CAM.

\noindent\textbf{Coherent Entity Candidate.} Entity candidates are constructed within each view by identifying representative centers and determining entity composition. To identify representative centers, ECO first defines neighborhoods according to spatial affinity and feature compatibility in each view. For the environment view, the neighborhood relations $\mathcal{N}^{E}$ can be defined as follows.
\begin{equation}
\mathcal{N}^{E}
=
\left\{
(i,j) \mid D_{ij}^{EE} \leq \tau_{E},\;
S_{ij}^{EE} \geq \eta
\right\},
\label{eq:intra_view_neighborhood}
\end{equation}
where $\tau_{E}$ and $\eta$ specify the maximum spatial separation and minimum feature similarity. Based on $\mathcal{N}^{E}$, ECO iteratively selects the uncovered location with the largest number of uncovered neighbors as a representative entity center, then marks the center and neighbors as covered. This process continues until all locations are covered, yielding a set of representative centers $\mathcal{C}^{E}$.

Subsequently, given a representative entity center at location $k \in \mathcal{C}^{E}$ , the entity composition $\mathcal{E}_{k}^{E}$ is determined as follows.
\begin{equation}
\mathcal{E}_{k}^{E}
=
\left\{
\mathbf{f}_{i}^{E}
\;\middle|\;
k =
\operatorname*{arg\,max}_{\substack{
j \in \mathcal{C}^{E} \\
(i,j) \in \mathcal{N}^{E}
}}
\left[
-\lambda(D_{ij}^{EE})^{2} + S_{ij}^{EE}
\right]
\right\}.
\label{eq:entity_composition}
\end{equation}
The resulting entity candidates $\mathcal{E}^{E} = \{\mathcal{E}_{k}^{E}\}_{k \in \mathcal{C}^{E}}$ are constructed around all representative centers, and the wrist view entity candidates $\mathcal{E}^{W} = \{\mathcal{E}_{k}^{W}\}_{k \in \mathcal{C}^{W}}$ are obtained through a similar procedure.

\noindent\textbf{Cross-View Entity Organization.} In order to address cross-view information fragmentation, ECO establishes relations between entity candidates across views by aggregating the cross-view associations, $\mathcal{A}^{EW}$ and $\mathcal{A}^{WE}$, from CAM. For an environment view candidate $\mathcal{E}_{k}^{E} \in \mathcal{E}^{E}$ and a wrist view candidate $\mathcal{E}_{l}^{W} \in \mathcal{E}^{W}$, the entity pair relations can be quantified as follows.
\begin{equation}
R_{kl}
=
\frac{
\sum\limits_{\mathbf{f}_{i}^{E} \in \mathcal{E}_{k}^{E}}
\sum\limits_{\mathbf{f}_{j}^{W} \in \mathcal{E}_{l}^{W}}
\mathcal{A}_{ij}^{EW}
}{
2\lvert \mathcal{E}_{k}^{E} \rvert
}
+
\frac{
\sum\limits_{\mathbf{f}_{i}^{W} \in \mathcal{E}_{l}^{W}}
\sum\limits_{\mathbf{f}_{j}^{E} \in \mathcal{E}_{k}^{E}}
\mathcal{A}_{ij}^{WE}
}{
2\lvert \mathcal{E}_{l}^{W} \rvert
},
\label{eq:entity_pair_relation}
\end{equation}
where $\lvert \cdot \rvert$ denotes set cardinality. To avoid inconsistent cross-view association, candidate entity pairs are organized by prioritizing higher $R_{kl}$, with each entity exclusively paired. The resulting pairs form the candidate entity pair set $\mathcal{E}^{EW}$.


By organizing local 3D features into entity candidates and modeling cross-view relations, ECO retains 3D evidence within cross-view entity pairs, thereby maintaining entity coherence and addressing cross-view information fragmentation.

\subsection{Consistent Action Conditioning}
\label{sec:method_cac}

To condition action prediction on cross-view coherent 3D entity evidence, CAC retrieves task-relevant entity pairs from $\mathcal{E}^{EW}$ and supplies the action expert with the corresponding 3D features.

\noindent\textbf{Task-Relevant Entity Retrieval.} In order to guide entity retrieval with task semantics, CAC employs a learnable query to attend to the vision-language feature from the final VLM layer, resulting in a task query $\mathbf{F}^{\mathrm{Task}}$. For an entity candidate $\mathcal{E}_{k}^{E}$, the task relevance is computed as follows.
\begin{equation}
T_{k}^{E}
=
\frac{1}{\lvert \mathcal{E}_{k}^{E} \rvert}
\sum_{\mathbf{f}_{i}^{E} \in \mathcal{E}_{k}^{E}}
\left\langle
\mathbf{F}^{\mathrm{Task}}, \widetilde{\mathbf{f}}_{i}^{E}
\right\rangle,
\label{eq:entity_task_relevance}
\end{equation}
where $\widetilde{\mathbf{f}}_{i}^{E}$ denotes the projected 3D feature from the entity candidate $\mathcal{E}_{k}^{E}$, and $T_{k}^{E}$ denotes the task relevance of the entity candidate. The same procedure yields $T_{l}^{W}$ for the corresponding wrist view candidate $\mathcal{E}_{l}^{W}$ paired with $\mathcal{E}_{k}^{E}$. For the candidate entity pair $(\mathcal{E}_{k}^{E}, \mathcal{E}_{l}^{W})$, the retrieval score is computed by averaging $T_{k}^{E}$ and $T_{l}^{W}$. CAC iteratively retrieves the entity pair with the highest retrieval score until $K$ pairs are retrieved or no candidate pair remains. The 3D entity features $\mathbf{F}^{EW}$ from both views of the retrieved pairs serve as conditioning inputs for action prediction.

\noindent\textbf{Conditioning Action Prediction.} With task-relevant entities identified, CAC incorporates the corresponding 3D features into action prediction. In the $n$-th layer of the action expert, the action feature attends to the vision-language features from the corresponding VLM layer, yielding $\mathbf{Q}_{n}^{A}$. The subsequent conditioning interaction with $\mathbf{F}^{EW}$ can be formulated as follows.
\begin{equation}
\widetilde{\mathbf{Q}}_{n}^{A}
=
\mathbf{Q}_{n}^{A}
+
\operatorname{Cross\text{-}Attn}_{n}
\left(
\mathbf{Q}_{n}^{A}, \mathbf{F}^{EW}, \mathbf{F}^{EW}; \mathbf{g}
\right),
\label{eq:action_conditioning}
\end{equation}
where $\widetilde{\mathbf{Q}}_{n}^{A}$ denotes the conditioned action feature propagated through the action expert for further prediction. For the retrieved entity pair $(\mathcal{E}_k^E,\mathcal{E}_l^W)$, $\mathbf{g}$ contains the associated gate $g_{kl}=\sigma\!\left(R_{kl}+(T_k^E+T_l^W)/2\right)$ to weight the corresponding 3D entity evidence according to task relevance and cross-view relation. 

\subsection{Training and Inference}
\label{sec:method_other}

\noindent\textbf{Training.} ConsisX integrates the proposed CAM, ECO, and CAC modules with a pre-trained VLM, an action expert, and a pre-trained 3D vision model. During training, the parameters of the action expert and the proposed modules are optimized, with the pre-trained VLM and 3D vision model kept frozen. All of the trainable parameters are jointly updated using the flow-matching objective of the base policy \cite{intelligence2025pi_}.

\noindent\textbf{Inference.} ConsisX takes the multi-view observations and language instruction as input and employs the VLM for joint encoding and the 3D vision model for 3D feature extraction. Subsequently, CAM establishes cross-view association to support ECO in maintaining entity coherence, enabling CAC to retrieve task-relevant 3D entity features. Conditioned on 3D entity evidence, the action expert iteratively transforms Gaussian noise into an action chunk.


\section{Experiments}
\label{sec:experiments}

\begin{table*}[t]
\centering
\caption{Comparison on the LIBERO benchmark. The best results are shown in \textbf{bold}, and the second best results are \underline{underlined}. Avg. denotes the mean of the success rates for the 4 task suites.}
\label{tab:libero}
\small
\setlength{\tabcolsep}{13pt}
\begin{tabular}{
@{}
>{\columncolor{white}[0pt][\tabcolsep]}l
|c|c|c|c|c|c
}
\toprule
\textbf{Method} & \textbf{Publication} & \textbf{Spatial} & \textbf{Object} & \textbf{Goal} & \textbf{Long} & \textbf{Avg.} \\
\midrule
Octo \cite{team2024octo} & RSS 2024 & 78.9 & 85.7 & 84.6 & 51.1 & 75.1 \\
OpenVLA \cite{kim2024openvla} & CoRL 2024 & 84.7 & 88.4 & 79.2 & 53.7 & 76.5 \\
TraceVLA \cite{zheng2025tracevla} & ICLR 2025 & 84.6 & 85.2 & 75.1 & 54.1 & 74.8 \\
CoT-VLA \cite{zhao2025cot} & CVPR 2025 & 87.5 & 91.6 & 87.6 & 69.0 & 83.9 \\
SpatialVLA \cite{qu2025spatialvla} & RSS 2025 & 88.2 & 89.9 & 78.6 & 55.5 & 78.1 \\
4D-VLA \cite{zhang20264d} & NeurIPS 2025 & 88.9 & 95.2 & 90.9 & 79.1 & 88.6 \\
\piZero-FAST \cite{pertsch2025fast} & RSS 2025 & 96.4 & 96.8 & 88.6 & 60.2 & 85.5 \\
\piZero \cite{black2024pi_0} & RSS 2025 & 96.8 & 98.8 & 95.8 & 85.2 & 94.2 \\
UniVLA \cite{bu2025univla} & RSS 2025 & 96.5 & 96.8 & 95.6 & 92.0 & 95.2 \\
OpenVLA-OFT \cite{kim2025fine} & RSS 2025 & 97.6 & 98.4 & 97.9 & 94.5 & 97.1 \\
\piHalf \cite{intelligence2025pi_} & CoRL 2025 & \second{98.8} & 98.2 & 98.0 & 92.4 & 96.9 \\
VLA-Adapter \cite{wang2026vla} & AAAI 2026 & 97.8 & 99.2 & 97.2 & 95.0 & 97.3 \\
MemoryVLA \cite{shi2026memoryvla} & ICLR 2026 & 98.4 & 98.4 & 96.4 & 93.4 & 96.7 \\
X-VLA \cite{zheng2026x} & ICLR 2026 & 98.2 & 98.6 & 97.8 & \best{97.6} & 98.1 \\
VIPA-VLA \cite{feng2026spatial} & CVPR 2026 & 96.6 & 98.6 & 97.0 & 95.0 & 96.8 \\
GeoPredict \cite{qian2026geopredict} & CVPR 2026 & 98.0 & 98.2 & 95.7 & 94.0 & 96.5 \\
ConsisVLA-4D \cite{li2026consisvla} & CVPR 2026 & \second{98.8} & \best{99.8} & 98.0 & 95.6 & 98.1 \\
ACoT-VLA \cite{zhong2026acot} & CVPR 2026 & 98.6 & 99.0 & \best{99.4} & 97.0 & \second{98.5} \\
\midrule
\rowcolor{oursbg}
\textbf{ConsisX} & Ours & \best{99.4} & \second{99.6} & \second{99.0} & \second{97.2} & \best{98.8} \\
\bottomrule
\end{tabular}
\end{table*}

We introduce the experimental setup in \cref{sec:experimental_setup}, evaluate ConsisX through simulation experiments in \cref{sec:simulation_results}, ablation studies in \cref{sec:ablation}, and real-world experiments in \cref{sec:real_world_results}, and conclude with qualitative analyses in \cref{sec:qualitative}.

\subsection{Experimental Setup}
\label{sec:experimental_setup}

\textbf{Simulation benchmarks.}
ConsisX is evaluated on 4 simulation benchmarks as detailed below.

\begin{itemize}[leftmargin=*, itemsep=2pt, topsep=0pt, parsep=0pt]
    \item \textbf{LIBERO.}
    LIBERO~\cite{liu2023libero} organizes manipulation tasks into 4 suites, \textit{i.e.}, LIBERO-Spatial (Spatial), LIBERO-Object (Object), LIBERO-Goal (Goal), and LIBERO-Long (Long), with 10 tasks per suite. For each task, the policy is trained on 50 human-teleoperated demonstrations. Evaluation involves 50 trials per task, totaling 2,000 trials across the 4 suites. The success rate for each suite and the mean across all 4 suites are reported.

    \item \textbf{LIBERO-Plus.}
    LIBERO-Plus~\cite{fei2025libero} builds on LIBERO with 10,030 episodes covering 7 perturbation types, \textit{i.e.}, camera viewpoint (Camera), robot initial state (Robot), language instruction (Language), lighting condition (Light), background texture (Background), sensor noise (Noise), and object layout (Layout). For evaluation, the policies optimized on LIBERO are directly tested on LIBERO-Plus. The success rate for each perturbation type and the overall result are reported.

    \item \textbf{CALVIN.}
    CALVIN~\cite{mees2022calvin} provides 34 manipulation tasks in 4 environments, A, B, C, and D. The ABC$\rightarrow$D protocol optimizes policies in A, B, and C, followed by evaluation in D with 1,000 sequences of 5 tasks each. The average successful task sequence length is reported. The detailed cumulative success rates are provided in the appendix.
    
    \item \textbf{LIBERO-PRO.}
    LIBERO-PRO~\cite{zhou2025libero} evaluates policies on 4 LIBERO task suites under perturbations including object attributes, initial spatial configurations, semantic instruction variations, and task goals. The policies optimized on LIBERO are directly tested on LIBERO-PRO without additional training. The overall success rate is reported. The detailed results are provided in the appendix.
\end{itemize}

\begin{table*}[t]
\centering
\caption{Comparison on the LIBERO-Plus benchmark. The best results are shown in \textbf{bold}, and the second best results are \underline{underlined}.}
\label{tab:libero_plus}
\small
\setlength{\tabcolsep}{6pt}
\begin{tabular}{
@{}
>{\columncolor{white}[0pt][\tabcolsep]}l
|c|c|c|c|c|c|c|c|c
}
\toprule
\textbf{Method} & \textbf{Publication} & \textbf{Camera} & \textbf{Robot} & \textbf{Language} & \textbf{Light} & \textbf{Background} & \textbf{Noise} & \textbf{Layout} & \textbf{Overall} \\
\midrule

OpenVLA \cite{kim2024openvla} & CoRL 2024
& 0.8 & 3.5 & 23.0 & 8.1 & 34.8 & 15.2 & 28.5 & 15.6 \\

UniVLA \cite{bu2025univla} & RSS 2025
& 1.8 & 46.2 & 69.6 & 69.0 & 81.0 & 21.2 & 31.9 & 42.9 \\

\piZero-FAST \cite{pertsch2025fast} & RSS 2025
& 65.1 & 21.6 & 61.0 & 73.2 & 73.2 & 74.4 & 68.8 & 61.6 \\

OpenVLA-OFT \cite{kim2025fine} & RSS 2025
& 56.4 & 31.9 & 79.5 & 88.7 & 93.3 & 75.8 & 74.2 & 69.6 \\

\piZero \cite{black2024pi_0} & RSS 2025
& 61.0 & 40.8 & 63.5 & 89.3 & 84.1 & 80.1 & 76.4 & 69.4 \\

\piHalf \cite{intelligence2025pi_} & CoRL 2025
& \second{75.8} & 79.4 & 83.3 & 95.5
& 95.0 & \second{89.6} & 87.0 & 85.7 \\

DreamVLA \cite{zhang2026dreamvla} & NeurIPS 2025
& 65.0 & 40.9 & 63.5 & 85.7 & 82.7 & 85.0 & 74.0 & 69.9 \\

GuidedVLA \cite{jia2026guidedvla} & RSS 2026
& 73.7 & 51.4 & 62.6 & 94.6 & 89.0 & 85.2 & 79.9 & 75.4 \\

Libra-VLA \cite{wei2026libra} & ACL 2026
& 68.9 & 48.8 & \best{92.7} & \best{97.9}
& 93.4 & 86.3 & 77.5 & 79.5 \\

VLA-Adapter \cite{wang2026vla} & AAAI 2026
& 36.2 & 37.9 & 74.6 & 70.6 & 76.1 & 58.0 & 69.7 & 59.1 \\

AVA-VLA \cite{xiao2026ava} & CVPR 2026
& 55.5 & 25.9 & 85.6 & 95.5 & 88.9 & 78.0 & 74.1 & 70.1 \\


ACoT-VLA \cite{zhong2026acot} & CVPR 2026
& 72.6 & \best{82.6} & 87.5 & \second{97.7}
& \second{96.5} & 87.8 & \second{88.1} & \second{86.6} \\

ResVLA \cite{zhong2026noise} & ICML 2026
& 53.2 & 57.5 & 88.2 & 94.5 & 96.3 & 81.8 & 78.3 & 76.9 \\



\midrule
\rowcolor{oursbg}
\textbf{ConsisX} & Ours
& \best{78.2}
& \second{81.2}
& \second{88.4}
& 96.7
& \best{96.9}
& \best{90.8}
& \best{89.3}
& \best{88.0} \\
\bottomrule
\end{tabular}
\end{table*}

\begin{table}[t]
\centering
\caption{
Comparison on the CALVIN and LIBERO-PRO benchmarks.
The best results are shown in \textbf{bold}, and the second best results are \underline{underlined}.}
\label{tab:additional_benchmarks}
\small
\setlength{\tabcolsep}{2.8pt}

\begin{tabular}{l|c|c|c}
\toprule
\multirow{2}{*}{\textbf{Method}}
& \multirow{2}{*}{\textbf{Publication}}
& \textbf{CALVIN}
& \textbf{LIBERO-PRO} \\
& & \textbf{Length} & \textbf{Success Rate} \\
\midrule
OpenVLA \cite{kim2024openvla}
& CoRL 2024
& 3.27
& 47.6 \\
$\pi_{0}$ \cite{black2024pi_0}
& RSS 2025
& 3.84
& 45.3 \\
OpenVLA-OFT \cite{kim2025fine}
& RSS 2025
& 4.10
& 23.3 \\
$\pi_{0.5}$ \cite{intelligence2025pi_}
& CoRL 2025
& 4.31
& \second{53.3} \\
VLA-Adapter \cite{wang2026vla}
& AAAI 2026
& 4.42
& 46.1 \\
X-VLA \cite{zheng2026x}
& ICLR 2026
& \second{4.43}
& 44.8 \\
\midrule
\rowcolor{oursbg}
\textbf{ConsisX}
& Ours
& \best{4.47}
& \best{63.3} \\
\bottomrule
\end{tabular}
\end{table}

\noindent\textbf{Real-world experiments.}
ConsisX is evaluated with 3 real-world manipulation tasks on an AgileX Piper robot. For each task, policies are optimized with 50 human-teleoperated demonstrations and evaluated in 20 trials. A trial is considered successful only when the task goal is achieved. The success rates for the 3 tasks are reported.

\begin{itemize}[leftmargin=*, itemsep=2pt, topsep=0pt, parsep=0pt]
    \item \textbf{Cross-View Association.}
    Given cubes of the same color and varying sizes, the robot is instructed to place the cube of the specified size on a plate. The cubes share a similar appearance and exhibit size ambiguity across views, thereby evaluating the capability for cross-view association during robotic manipulation.

    \item \textbf{Cross-View Entity Coherence.}
    The robot is instructed to nest the paper cups. The environment and wrist cameras capture different local details of the same cup, thereby evaluating the capability to integrate complementary information across views for entity coherence perception.

    \item \textbf{General Manipulation.}
    The robot is instructed to place the fruit in a box and leave distractors untouched, thereby evaluating the general manipulation capability of the policies.
\end{itemize}

\noindent\textbf{Implementation details.}
Visual observations from multi-view cameras are resized to $224 \times 224$ for the VLM and $512 \times 512$ for the 3D vision model. Action chunks contain $H=10$ steps in simulation and $H=30$ steps in real-world experiments. Both the VLM and the action expert contain 18 layers, with hidden dimensions of $2048$ and $1024$, respectively. ConsisX adopts pre-trained VGGT-$\Omega$~\cite{wang2026vggt} as the 3D vision model. The 3D feature for each view is average-pooled to \(16 \times 16\) spatial resolution with \(N=256\).

\subsection{Simulation Results}
\label{sec:simulation_results}

\textbf{LIBERO.}
As shown in \Cref{tab:libero}, ConsisX ranks first with an average success rate of 98.8\%. 
Compared with the base policy, ConsisX improves in all 4 suites, with the largest gain of 4.8\% on Long. For the Spatial and Object suites, ConsisX achieves notable success rates of 99.4\% and 99.6\%, respectively, highlighting the benefits of cross-view consistency thinking in 3D for both spatial understanding and object manipulation.

\noindent\textbf{LIBERO-Plus.}
ConsisX achieves an overall success rate of 88.0\%, outperforming all other methods in \Cref{tab:libero_plus}. Across all perturbations, ConsisX makes consistent improvements over the base policy and records an overall gain of 2.3\%. For the camera viewpoint perturbation, ConsisX achieves the highest success rate of 78.2\%, exceeding the base policy by 2.4\% and supporting the proposed cross-view design for consistent task manipulation under viewpoint changes.

\noindent\textbf{CALVIN.}
In CALVIN, ConsisX performs manipulation tasks with an average completed sequence length of 4.47, improving on the base policy by 0.16 and attaining the highest result in \Cref{tab:additional_benchmarks}. The increased average sequence length supports the role of cross-view consistency in maintaining execution stability during long-horizon manipulation.

\noindent\textbf{LIBERO-PRO.}
As reported in \Cref{tab:additional_benchmarks}, ConsisX obtains an overall success rate of 63.3\% on LIBERO-PRO, surpassing the base policy by 10.0\%. The improvement further reflects the capability of ConsisX to maintain reliable robotic manipulation performance under diverse perturbations.

\begin{table}[t]
\centering
\caption{
Ablation of the 3D spatial constraint on the LIBERO benchmark.}
\label{tab:ablation_spatial}
\small
\setlength{\tabcolsep}{2.6pt}

\begin{tabular}{@{}cc|cccc|c@{}}
\toprule
\textbf{Cross-View}
& \textbf{Within-View}
& \textbf{Spatial}
& \textbf{Object}
& \textbf{Goal}
& \textbf{Long}
& \textbf{Avg.} \\
\midrule
\xmark & \xmark
& 99.0 & 99.4 & 98.6 & 95.6 & 98.2 \\
\cmark & \xmark
& \textbf{99.4} & 99.0 & 98.4 & 96.2 & 98.3 \\
\xmark & \cmark
& 99.2 & 99.2 & 98.8 & 96.8 & 98.5 \\
\cmark & \cmark
& \textbf{99.4}
& \textbf{99.6}
& \textbf{99.0}
& \textbf{97.2}
& \textbf{98.8} \\
\bottomrule
\end{tabular}
\end{table}

\begin{table}[t]
\centering
\caption{
Ablation of the entity coherency modeling on the LIBERO benchmark.}
\label{tab:ablation_entity}
\small
\setlength{\tabcolsep}{3pt}

\begin{tabular}{@{}cc|cccc|c@{}}
\toprule
\textbf{Entity}
& \textbf{Task}
& \multirow{2}{*}{\textbf{Spatial}}
& \multirow{2}{*}{\textbf{Object}}
& \multirow{2}{*}{\textbf{Goal}}
& \multirow{2}{*}{\textbf{Long}}
& \multirow{2}{*}{\textbf{Avg.}} \\
\textbf{Organization}
& \textbf{Relevance}
& & & & & \\
\midrule
\xmark & \xmark
& 99.0 & 98.4 & 98.2 & 94.8 & 97.6 \\
\xmark & \cmark
& 98.8 & 99.0 & \textbf{99.0} & 95.6 & 98.1 \\
\cmark & \cmark
& \textbf{99.4}
& \textbf{99.6}
& \textbf{99.0}
& \textbf{97.2}
& \textbf{98.8} \\
\bottomrule
\end{tabular}
\end{table}

\begin{figure}[t]
    \centering
    \includegraphics[width=\columnwidth]{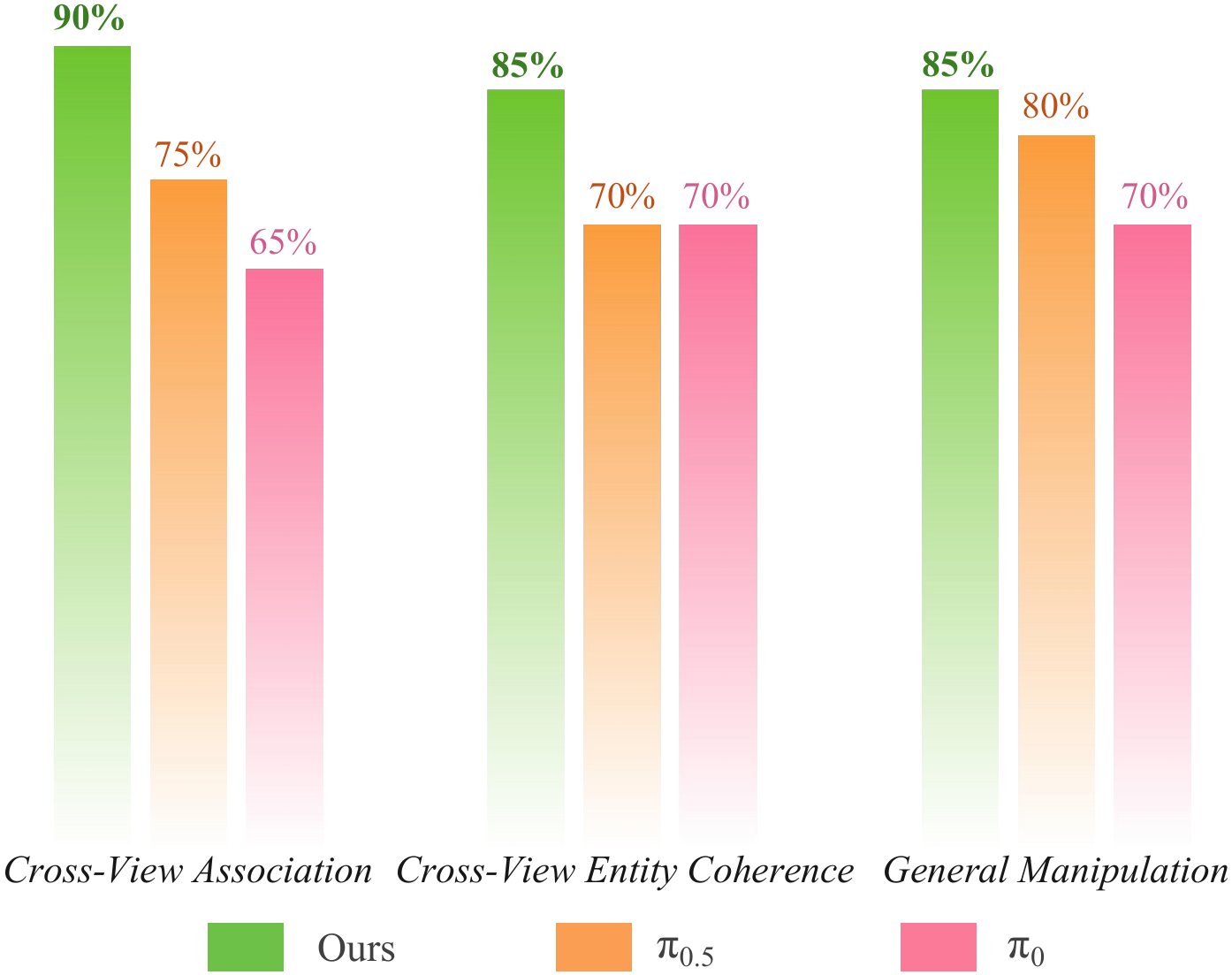}
    \caption{Comparison on the real-world experiments.}
    \label{fig:real-world}
\end{figure}

\begin{figure}[t]
    \centering
    \includegraphics[width=\columnwidth]{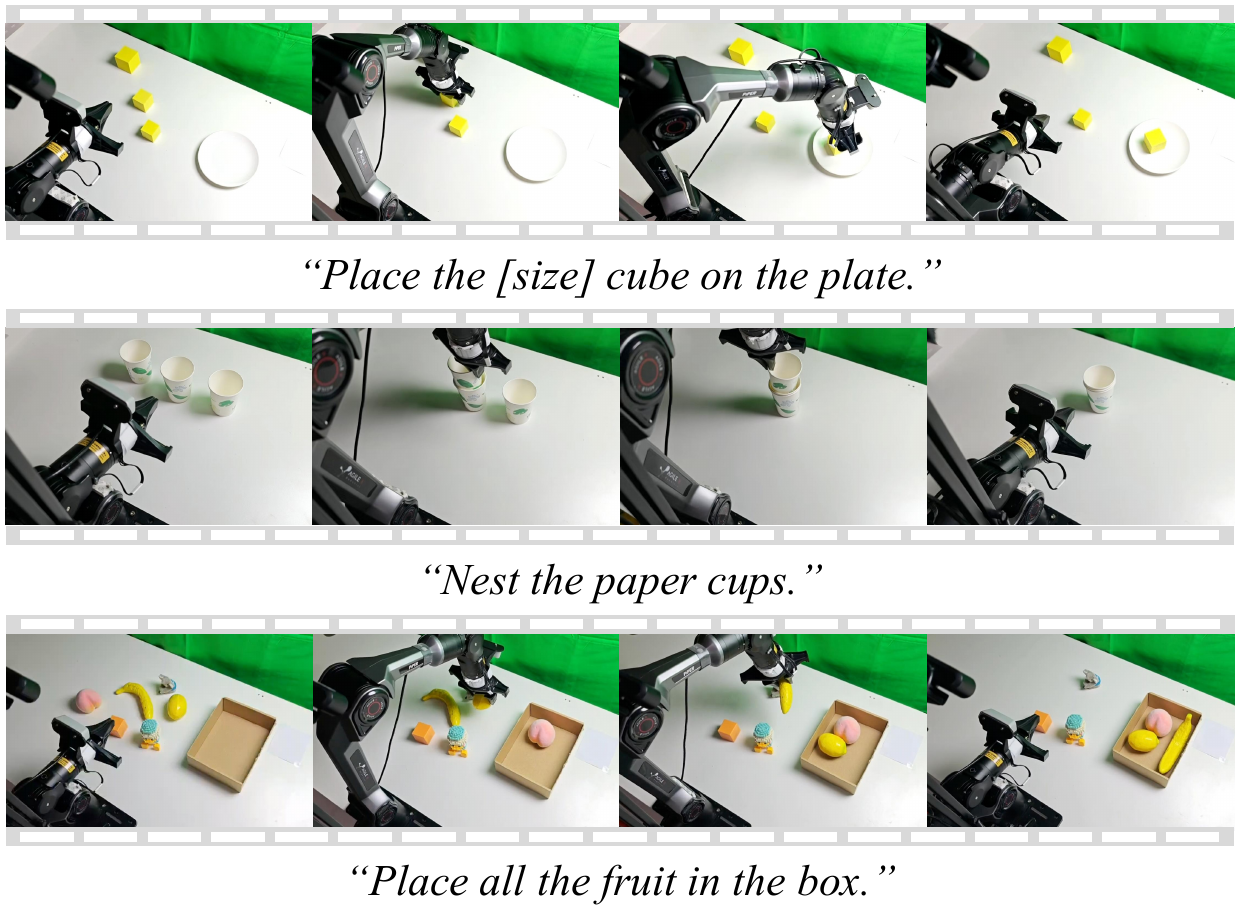}
    \caption{Visualization of the manipulation tasks in the real world.}
    \label{fig:real-visualization}
\end{figure}

\begin{figure}[t]
    \centering
    \includegraphics[width=\columnwidth]{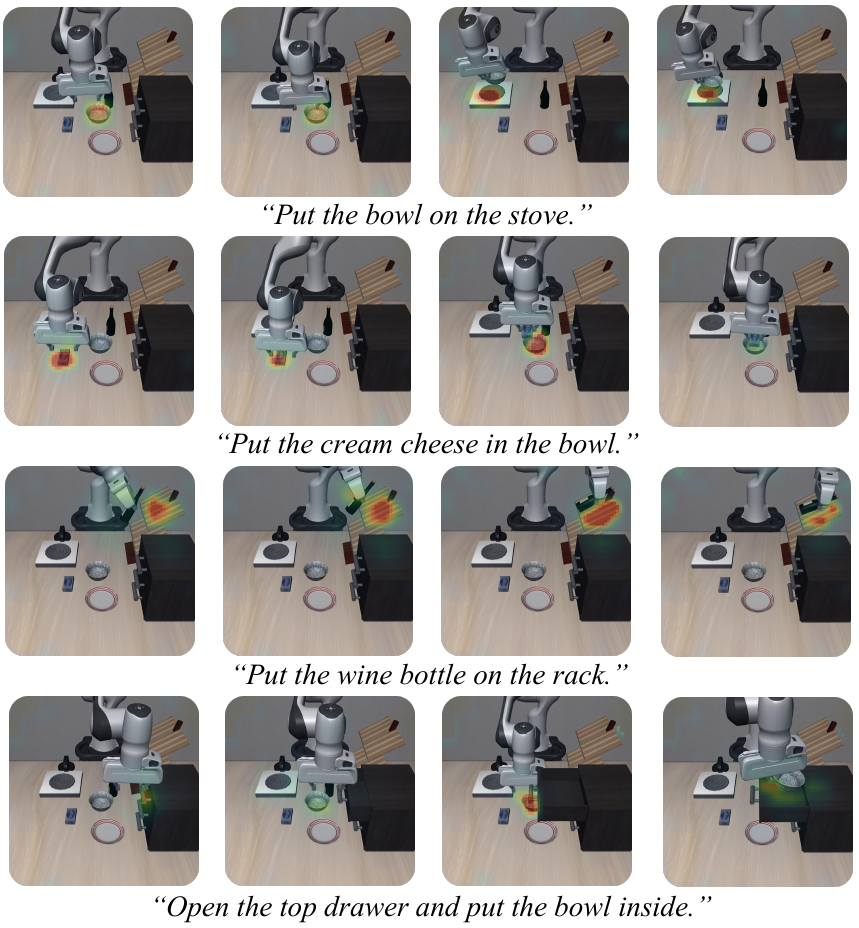}
    \caption{Visualization of action-entity attention maps in CAC.}
    \label{fig:qualitative}
\end{figure}

\subsection{Ablation Study}
\label{sec:ablation}

The ablation studies are conducted on the LIBERO benchmark, focusing on the roles of 3D spatial constraint and coherent entity modeling in addressing cross-view incorrect association and information fragmentation, respectively. The results are reported in \Cref{tab:ablation_spatial} and \Cref{tab:ablation_entity}.

\noindent\textbf{Effect of 3D Spatial Constraint.} As shown in \Cref{tab:ablation_spatial}, we compare the effect of 3D spatial constraint in cross-view association and within-view entity organization. The method in the first row performs cross-view association and within-view entity organization relying on feature similarity without 3D spatial constraint. In the second row, 3D spatial constraint is introduced for cross-view association, raising the average success rate to 98.3\%. The third row restricts spatial constraints to within-view entity organization, resulting in 98.5\%. Combining both constraints in the fourth row achieves an overall gain of 0.6\%, supporting the effectiveness of 3D spatial constraint for task manipulation.

\noindent\textbf{Analysis of Entity Coherency Modeling.} As shown in \Cref{tab:ablation_entity}, we analyze the design of entity coherency modeling. In the first row, the action expert is conditioned on all 3D features extracted from both views, achieving an average success rate of 97.6\%. The method in the second row focuses on the top task-relevant 3D features, resulting in 98.1\%. Further incorporating cross-view entity organization, method in the third row provides a cumulative gain of 1.2\%, reflecting the value of organizing cross-view fragmented information into coherent 3D entity evidence.

\subsection{Real-World Experiments}
\label{sec:real_world_results}
As shown in \cref{fig:real-world}, ConsisX achieves the highest success rate on each of the 3 real-world manipulation tasks. For Cross-View Association, ConsisX improves on the base policy by 15\%, showing the mitigation of association ambiguity across views. In Cross-View Entity Coherence, ConsisX raises the success rate from 70\% to 85\%, demonstrating the ability to integrate fragmented cross-view information. ConsisX also achieves 85\% on General Manipulation, showing reliable general manipulation capability. The visualizations are presented in \cref{fig:real-visualization}.

\subsection{Qualitative Analysis.}
\label{sec:qualitative}
To illustrate how coherent 3D entity evidence conditions action prediction, we visualize the action-entity attention maps of CAC in \cref{fig:qualitative}. As execution progresses, the responses focus on task-relevant entities and interaction regions, highlighting the target during grasping and the destination during placement. The shift attention response suggests that action prediction attends to the 3D evidence from task-relevant entities at different stages of manipulation, providing qualitative results for cross-view consistency thinking in 3D.

\section{Conclusion}
\label{sec:conclusion} 
Existing VLA models rely on 2D feature similarity within VLMs to model cross-view interactions, which can lead to physical inconsistency in the form of incorrect cross-view association and cross-view information fragmentation. To address these challenges, we propose ConsisX, a novel VLA framework that integrates cross-view association and entity coherence to enable cross-view consistency thinking in 3D. ConsisX implements the consistency thinking following a progressive process, with CAM establishing spatially constrained cross-view associations, ECO organizing local 3D features into cross-view coherent entity pairs, and CAC conditioning action prediction on task-relevant 3D entity evidence. Evaluations on 4 simulation benchmarks and 3 real-world tasks highlight the value of cross-view consistency thinking in 3D for reliable robotic manipulation.

\section*{LLM Usage Statement}
ChatGPT assisted with manuscript writing, searches for related work, code debugging, and data organization. It also offered suggestions on figure presentation and layout.
{
    \small
    \bibliographystyle{ieeenat_fullname}
    \bibliography{main}
}


\end{document}